\documentclass[11pt,a4paper]{article}
\usepackage[hyperref]{emnlp-ijcnlp-2019}
\usepackage{times}
\usepackage{latexsym}

\usepackage{url}

\usepackage{times}
\usepackage{latexsym}

\usepackage{bm}
\usepackage{graphicx}

\usepackage{booktabs}
\usepackage{amsfonts}
\usepackage{setspace}
\usepackage{soul}
\usepackage{enumitem}

\usepackage{soulpos}
\usepackage{url}
\usepackage{adjustbox}
\usepackage{subcaption}
\usepackage{ragged2e}

\usepackage{comment}
\usepackage{multirow}
\usepackage{multicol}
\usepackage{booktabs}
\usepackage{array}
\usepackage{tikz}
\usepackage{pgfplots}
\pgfplotsset{width=5cm,compat=1.3, legend style={at={(0.40,0.40)},anchor=north west}}
\newlength\myindent
\usepackage{float} % alg float
\usepackage{amsmath}
\usepackage{algorithm,algorithmic}

\DeclareMathOperator*{\argmax}{argmax}
\DeclareMathOperator*{\argmin}{argmin}
\usepackage{xcolor}
\usepackage{colortbl}
\usepackage{color}
\definecolor{myorange}{RGB}{193, 128, 0}
\definecolor{myblue}{RGB}{5, 96, 175}
\definecolor{mygreen}{RGB}{34,139,34}

\definecolor{grey}{rgb}{0.8,0.8,0.8}

\usepackage[colorinlistoftodos]{todonotes}
\aclfinalcopy % Uncomment this line for the final submission
\title{
IMaT: Unsupervised 
Text 
Attribute % Rewriting
Transfer via Iterative \\ Matching and Translation
}

\author{Zhijing Jin\thanks{ {} {}  Equal Contribution} \\
  University of Hong Kong \\
  \texttt{zhijing.jin@connect.hku.hk} \\\And
  Di Jin$^{*}$ \\
  MIT \\
  \texttt{jindi15@mit.edu} \\\AND
  Jonas Mueller \\
  Amazon Web Services \\
  \texttt{jonasmue@amazon.com} \\\And
  Nicholas Matthews \\
  ASAPP Inc. \\
  \texttt{nmatthews@asapp.com} \\
  \\
  \And
  Enrico Santus \\
  MIT \\
  \texttt{esantus@csail.mit.edu} \\}

\date{}

\begin{document}
\maketitle
\begin{abstract}
 Text attribute transfer aims to automatically rewrite sentences such that they possess certain linguistic attributes, while simultaneously preserving their semantic content. This task remains challenging due to a  lack of supervised parallel data. Existing approaches try to explicitly disentangle content and attribute information, but this is difficult and often results in poor content-preservation and ungrammaticality. In contrast, we propose a simpler approach, \textbf{\underline{I}}terative \textbf{\underline{M}}atching \textbf{\underline{a}}nd \textbf{\underline{T}}ranslation (\textbf{IMaT}), which: (1) constructs a pseudo-parallel corpus by aligning a subset of semantically similar sentences from the source and the target corpora; (2) applies a standard sequence-to-sequence model to learn the attribute transfer; (3) iteratively improves the learned transfer function by refining imperfections in the alignment. In sentiment modification and formality transfer tasks, our  method outperforms complex state-of-the-art systems by a large margin. As an auxiliary contribution, we produce a publicly-available test set with human-generated transfer references.\footnote{The enriched test set is available at   \url{https://github.com/zhijing-jin/IMaT}}.

  \iffalse % TODO: include this in camera ready. Left out for anonymity
  \footnote{Code and data available at:  github.com/zhijing-jin/IMT}
  \fi
\end{abstract}

\section{Introduction}
\label{sec:intro}

% \begin{figure}[t]
%     \small
%     \centering
%     \includegraphics[width=0.45 \textwidth]{img/approach_overview.png}
%     \caption{An overview of the proposed Iterative Matching and Translation (IMT).}
%     %(Step 1) We construct pseudo pairs through semantic matching of sentence embeddings. 
%     %(Step 2) We transfer the sentence by iteratively translating and matching.
    
%     \label{fig:overview}
% \end{figure}

An intelligent natural language generation (NLG) system should be able to flexibly control linguistic attributes of the text it generates.  For instance, a certain level of formality should be maintained in serious situations, while informal conversations can be improved through a more relaxed style. The ability to generate or rewrite a certain text with some attributes controlled or transferred to meet pragmatic goals is widely needed in applications such as dialogue generation \cite{Oraby2018ControllingPS}, author obfuscation~\cite{Kacmarcik2006ObfuscatingDS,Juola2011AnalyzingSA}, and written communication assistance~\cite{Pavlick2016AnEA}. One typical example of a text attribute is the linguistic style, which refers to features of lexicons, syntax and phonology that collectively contribute to the identifiability of an instance of language \cite{gatt2018survey}.
% is originated from language variation such as accent, lexicon, morphology and syntax. 
For instance, given text written in an older style like the Shakespearean genre, a system may be tasked to convert it into modern-day colloquial English
% for example, that of simple English Wikipedia
\cite{Kabbara2016StylisticTI}. 
% ``Style transfer'' can be thus viewed as a  particular type of attribute-controlled text rewriting \cite{lample2018multipleattribute}.
% This ability can also help to synthesize style-specific data that can be used to complement imbalanced or biased datasets. Consider for example, datasets of user reviews, in which the majority of comments tend to be positive. Training classifiers on such data may lead to an unfair system, biased towards the majority class. Adding synthetic negative reviews to the training set could instead improve model fairness.

\begin{table}[b]
\centering
% \small
% \scriptsize
\footnotesize
\resizebox{\columnwidth}{!}{
\begin{tabular}{m{.5cm}l}
\toprule
\multicolumn{2}{c}{\textbf{{\color{mygreen}Positive} Sentiment $\leftrightarrow$ {\color{red}Negative} Sentiment}} \\
\hline
\textbf{Input} & {\color{mygreen}I love this place}, the service is {\color{mygreen}always great!}	 \\
\hline
CA&I know this place, the food is just a horrible!\\
MD &  I love this place, the service is always great!  \\
DAR &I did not like the homework of lasagna, not like it, .	  \\
\textbf{Ours} &I used to love this place, but the service is horrible now.	  \\
\toprule
\end{tabular}
}
\caption{Comparing outputs of transfer models. Our approach both transfers sentiment and preserves content. \textit{CrossAlignment} (CA) by \citet{shen2017style} loses content, \textit{MultiDecoder} (MD) by \citet{fu2017style} fails to modify  sentiment, and \textit{DeleteAndRetrieve} (DAR) by \citet{li2018delete} produces ungrammatical output. }
\label{table:example1}
\end{table}

% Such motivations have led to the formulation of the style transfer task, which aims to rewrite  a given input sentence in a new style while simultaneously preserving the content. Style here may refer to a broad range of linguistic phenomena, such as syntactic simplification, word substitution, or sentiment manipulation.
In general, a text attribute transfer system must be able to:
% needs to satisfy the following requirements: 
(1) produce sentences that conform to the target attribute, (2) preserve the source sentence's content, and (3) generate fluent language. Satisfying these requirements is challenging due to the typical absence of supervised parallel corpora exemplifying the desired attribute transformations.
With access to only non-parallel data, most existing approaches aim to tease apart content and attribute in a latent sentence representation. For instance, \citet{shen2017style} and \citet{fu2017style} utilize Generative Adversarial Networks (GANs) to learn this separation. Although these elaborate models can produce outputs that conform to the target attribute (requirement 1), their sentences are fraught with errors in terms of both content preservation and linguistic fluency. This is exemplified in Table~\ref{table:example1}, where the model of \citet{shen2017style} changes key content words from ``the service'' to the unrelated ``the food''. 
%Referring to the example provided in the first row of Table~\ref{table:example1}, the key content changes from ``green chile chimis'' to an unrelated object ``green \textit{crab dip}''.
In an effort to avoid the issues of GAN/autoencoder approaches, \citet{li2018delete} recently demonstrated that direct implementation of  heuristic transformations--such as modifying high polarity words in reviews for sentiment transfer-- can produce better results. Their work suggests attribute transfer can be addressed through simpler methods that avoid attempting to disentangle attribute and content information at the representation level. However, the proposed heuristic transformations are a bit too simple, relying on rule-based sentence reconstruction that often produces linguistically unnatural or ungrammatical outputs. Even worse, some unrelated words happen to be wrongly inserted by approach of \citet{li2018delete}, dramatically upsetting the sentence content as seen in Table~\ref{table:example1}. %, for example of ``They \textit{refuse to green chile chile chimis}'' as shown in Table~\ref{table:example1}.

In this paper, we propose the Iterative Matching and Translation (IMaT) framework which addresses the aforementioned limitations with regards to content inconsistency and ungrammaticality.
% Unlike previous methods, we do not explicitly aim for attribute and content disentanglement. Rather, inspired by the success of iterative back-translation approaches in machine \cite{artetxe2017unsupervised,lample2018phrase} and style translation \cite{zhang2018style,lample2018multipleattribute}, we propose 
% an iterative matching and translation framework.
%  , as it is shown in Figure \ref{fig:overview}. 
Our approach first constructs a pseudo-parallel corpus by \textbf{matching} a subset of semantically similar sentences from the source and the target corpora (which possess different attributes), then applies a standard sequence-to-sequence (Seq2Seq) \textbf{translation} model \cite{klein-etal-2017-opennmt} to learn the attribute transfer. We then use the translation results of the trained Seq2Seq model (from the previous iteration) to update the previously made pseudo-parallel corpus, so as to \textbf{refine} its quality. Such a matching-translation-refine procedure is iterated repeatedly until performance plateaus. The proposed methodology is simpler and more robust than previous GAN/autoencoder techniques, and is free of manual heuristics such as those used by \citet{li2018delete}.

We validate our Iterative Matching and Translation (IMaT) model in two attribute-controlled text rewriting tasks that aim to alter: the sentiment of \textsc{Yelp} reviews, and the formality of text in the \textsc{Formality} dataset. Both human and automatic evaluations suggest our method substantially outperforms alternative approaches \cite{shen2017style, fu2017style, li2018delete} in terms of: accuracy of attribute change, content preservation, and grammaticality. 
Our main contributions include: 

\begin{itemize}
    \item We propose a novel iterative matching and translation framework that is more straightforward than many existing approaches, by simply adapting a  sequence-to-sequence model to perform the attribute transfer.
    %  which refines the transfer model in iterations and largely improves the performance of unsupervised text attribute transfer. 
    % \item We treat the Seq2Seq component as a replaceable module, which can be updated with a more state-of-the-art model in the future, and thus making it adaptable to the rapid advances in Seq2Seq modeling \citep{dehghani2018universal}. 
    
    \item We achieve state-of-the-art performance on two text rewriting tasks involving sentiment modification task and  formality conversion.
    
    % by training our model under a pseudo-supervised setting.  Here, the retrieval of semantically similar sentences is combined with outputs generated from our translation model to together act as supervised training data for the ultimate task at hand.
    % \item Having bootstrapped a pseudo-parallel corpus, we can leverage a standard Seq2Seq translation model (with attention which enables rewriting long sentences) for the attribute transfer task. This enables our framework to naturally improve with rapid advancements in Seq2Seq modeling such as the universal transformer \citep{dehghani2018universal}. 
    
    % The availability of pseudo-parallel corpus enables the use of attention-based Seq2Seq model which directly models the whole attribute shift process, and the attention mechanism can better make the decision of preserving the content words and transferring the attribute-related words.
    
    % \item We propose a novel iterative matching and translation process, which improves the transfer model step by step, somewhat akin to Expectation-Maximization (EM).
    
    % Similar to the expectation-maximization (EM) algorithm based training strategy, the iterative matching and translation process can self-improve the model until good convergence.
    
    % \item We critically assessed the quality of five popular datasets for text attribute transfer.
    \item We release an additional set of 800 sentences rewritten by humans tasked with the sentiment transfer task
    % asked to perform the same sentiment modification task 
    for the \textsc{Yelp} review test set. This enables future researchers to evaluate more diverse transfer outputs.
    
    % \item We proposed a simple but effective approach to text attribute rewriting by iterative matching and translation.

\end{itemize}

\section{Related Work}
\label{sec:relatedwork}

Attribute-controlled text rewriting remains a long-standing problem in NLG, where most work has focused on studying the stylistic variation in text~\cite{gatt2018survey}. Early contributions in this area defined stylistic features using rules to vary generation \cite{brooke2010automatic}. For instance, \citet{sheikha2011generation} proposed an adaptation of the SimpleNLG realiser \cite{gatt2009data} to handle formal versus informal language via constructing lexicons of formality (e.g., are not vs. aren't). More contemporary approaches have tended to eschew rules in favour of data-driven methods to identify relevant linguistic features to stylistic attributes \cite{ballesteros2015data,di2008trainable,krahmer2012}. For example, Mairesse and Walker's PERSONAGE system \cite{mairesse2011controlling} uses machine-learning models to take as inputs a list of real-valued style parameters and generate sentences to project different personality traits. 

In the past few years, attribute-controlled NLG has witnessed renewed
interest by researchers working on neural approaches to generation~\cite{hu2017toward, jhamtani2017shakespearizing,  melnyk2017improved, mueller2017sequence, zhang2018style, prabhumoye2018style, niu2018polite}. 
% Much of the existing work follows style transfer methods initially developed for images~\cite{gatys2016image, zhu2017unpaired, liu2016coupled}.
% One popular technique for unsupervised image style transfer is the CycleGAN model proposed by \citet{zhu2017unpaired}, which performs image-to-image translation between domains in both directions via a pair of mappings trained to minimize a combination of adversarial and cycle loss functions.
Among them, many attribute-controlled text rewriting methods similarly employ GAN-based models to disentangle the content and style of text in a shared latent space \citep{shen2017style, fu2017style}.  However, existing work that applies these ideas to text suffers from both training difficulty~\cite{salimans2016improved, arjovsky2017towards, bousmalis2017unsupervised}, and ineffective manipulation of the latent space which leads to content loss~\cite{li2018delete} and generation of grammatically-incorrect sentences. Other lines of research avoid adversarial training altogether. \citet{li2018delete} proposed a much simpler approach: identify style-carrying n-grams, replace them with phrases of the opposite style, and train a neural language model to combine them in a natural way. Despite outperforming the adversarial approaches, its  performance is dependent on the availability of an accurate word identifier, a precise word replacement selector and a perfect language model to fix the grammatical errors introduced by the crude swap. 

% Experiments show that the generated sentences in this fashion have many grammatical errors, as reported in Table~\ref{table:example1} and  Section~\ref{sec_human_results}.

Recent work improves upon adversarial approaches by additionally leveraging the idea of back translation~\cite{santos2018fighting,logeswaran2018content,lample2018multipleattribute,prabhumoye2018style}. 
It was previously used for
% , like the cycle loss for images, leads to better latent representations of sentences.
% While CycleGAN shows qualitatively strong results for images~\cite{zhu2017unpaired}, the corresponding results for text style transfer are less impressive.  It remains challenging to apply the same adversarial/cycle-training on text, due to its discrete nature.
% This line of work offers limited increase of performance due to the bottleneck in unsupervised text generation from a latent space. \myworries{can you specify a bit better what this bottleneck is?}\jonascomment{I've removed this statement entirely; our seq2seq model is also generating text from a latent vector...}
% We propose a similarly simple approach to text style transfer, inspired by the incremental training of pseudo-parallel corpora in 
unsupervised Statistical Machine Translation (SMT)~\cite{fung1998ir,munteanu2004improved,smith2010extracting}
% This approach was further applied in recent years to the area of 
and Neural Machine Translation (NMT)~\citep{conneau2017word,lample2017unsupervised,artetxe2017unsupervised}, where it iteratively takes the pseudo pairs to train a translation model and then use the refined model to generate new pseudo-parallel pairs with enhanced quality. However, the success of this method relies on good quality of the pseudo-parallel pairs. Our approach proposes using retrieved sentences from the corpus based on semantic similarity as a decent starting point and then refining them using the trained translation models iteratively.

% Our approach, in the context of style transfer, further enforces content preservation between pseudo pairs by minimizing their Word Mover Distance~\cite{huang2016supervised}. 

% \myworries{I like both versions. The new one seems more mature and discoursive. Make sure that all info are present. Not sure if you want to say a word about what optimal transport is, but just a word.}
%\note{I am keeping the new version. Thx to both Enrico and Jonas for making the language more mature!}
%\note{infos in two versions are very much kept}
%\note{Maybe let's not include OT, as I saw Jonas deleting the word OT from this Related Work.}

\section{Method}
\label{sec:method}

%%%%% THIS IS INTERESTING, BUT SHOULD GO IN THE METHOD %%%%%
%In our iterative model, we gradually expand the sentence-pairs used to train $M^{(t)}$, therefore exposing $M^{(t)}$ to increasingly large portions of the $\mathbb{P}_{X_1},\mathbb{P}_{X_2}$, and encouraging the final map $M^{(T)}$ to reduce the distance between $M^{(T)}(\mathbb{P}_{X_1})$ and $\mathbb{P}_{X_2}$.

\begin{figure*}[t]
    \small
    \centering
    \includegraphics[width= \textwidth]{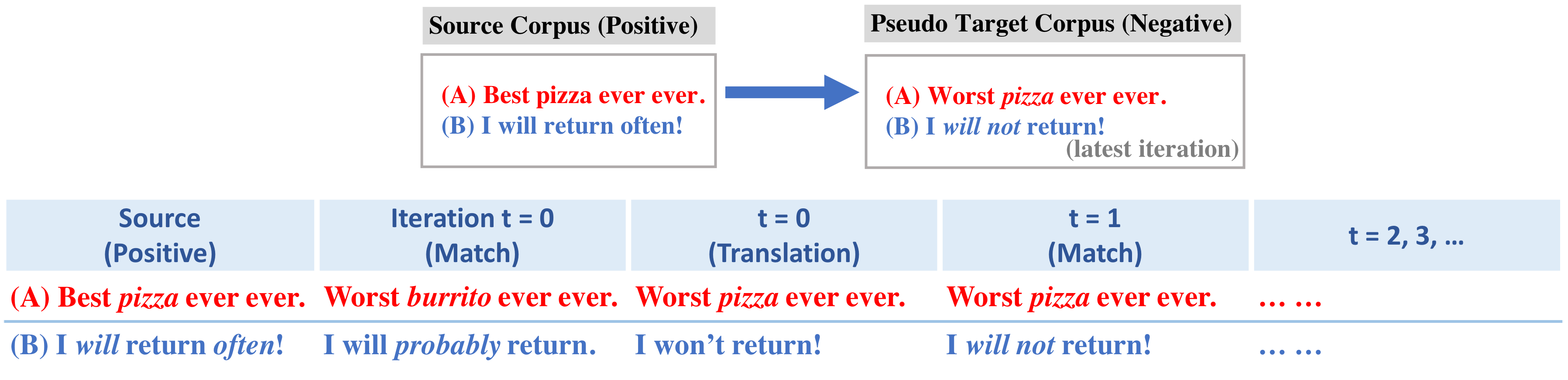}
    \caption{Iterative process of the algorithm to transfer the text style from positive to negative reviews. 
    %(Step 1) We construct pseudo pairs through semantic matching of sentence embeddings. 
    %(Step 2) We transfer the sentence by iteratively translating and matching.
    }
    \label{fig:method}
\end{figure*}

\subsection{Task Formulation}

Given two mono-style corpora $X=\{\bm{x_1},\cdots,\bm{x_n}\}$ with attribute $a_1$ and $Y=\{\bm{y_1},\cdots,\bm{y_m}\}$ with attribute $a_2$. Note that the alignment between $X$ and $Y$ corpora is not available. The unsupervised attribute-controlled rewriting task aims to learn a transformation $T^*(\cdot)$ from $X$ to $Y$ by optimizing the following objective:
% such that a source sentence $\bm{x_i}$ is rewritten to a sentence $y_i' = T$:
\begin{align}
    T^*(\cdot) = \argmin_{T(\cdot)}\ \|\bm{x}-T(\bm{x})\|, 
    \\
    \ s.t.\ \mathcal{A}(T(\bm{x}))=a_2,
\end{align}
where the norm $\|\cdot\|$ measures the content shift between two sentences, and $\mathcal{A}(\cdot)$ represents the attribute of a sentence. 
% where $\bm{\hat y_i}$ has the attribute $a_2$. Note that this transformation is to be learned in an unsupervised way.
Plainly put, a good attribute rewrite should ensure the attribute is changed to the desired value, and the content shift between the original sentence and rewrite is minimized.

% \paragraph{Style conversion objective}
% For the given two mono-style corpora $X$ and $Y$, we assume that every sentence conforms to the attribute distribution of the corpus. Therefore, for every original sentence $\bm{x_i} \in X, y_j \in Y$, it has the property that $\bm{x_i} \sim X$ and $y_j \sim Y$. Therefore, the rewrite $\bm{\hat y_i}$ need to carry the target attribute, namely
% \begin{align}\label{eq:subset_sq}
%     \bm{\hat y_i} \sim p(Y)
%     .
% \end{align}
% \paragraph{Minimal content change objective}
% If the attribute conversion is the only objective to achieve, then any sentence drawn from the original $Y$ can be considered as a success. Hence, another property of a successful attribute-controlled rewrite is the minimal content change. Namely, we need to make sure that the generated sentence preserves the essence of the original sentence and only manipulates the attribute. Since there exists no perfect way to represent the content similarity, we use Word Mover Distance of two sentences as an approximate. 

\subsection{Iterative Matching and Translation}

We propose an iterative matching and translation algorithm composed of the following three steps: 

\paragraph{(Step 1) Matching} 

In iteration $t=0$, 
% the whole iterative training scheme is started by the initial, unsupervised matching between the corpora $X$ and $Y$.
we construct a large pseudo-parallel corpus $\hat X$ and $\hat Y^{(0)}$ by pairing sentences from $X$ with those from $Y$. Specifically, we calculate the semantic cosine similarity score (detailed in Appendix~\ref{sec:appendix_exp_details}) between a sentence $\bm{x_i}\in X$ and every sentence $\bm{y}\in Y$, select the one with the highest score as $\bm{\hat y_i}$, and only keep the sentence pair if the similarity exceeds a threshold $\gamma$. $\hat X$ denotes the subset of the original corpus $X$ for which we find matches.

% We identify the pseudo target $\hat y$ by selecting the highest cosine similarity to the unsupervised sentence embedding of $x$. 
In later iterations $t\geq 1$, this matching process is different from that in the first iteration. We match the pseudo-parallel corpus $\hat X$ with $\hat Y^{(t)}$, which is refined in step 3 of the previous iteration, and obtain a \textit{temporary} matched corpus $\mathrm{Match}^{(t)}$. In this case, for any sentence $\bm{x_i}\in \hat X$, we have two pseudo-parallel sentences $\bm{\hat y_i}\in \hat Y^{(t)}$ and $\bm{\mathrm{match}_i}\in \mathrm{Match}^{(t)}$, both of which have the opposite attribute to $\bm{x_i}$. We then use Word Mover Distance (WMD), which will be detailed in Section~\ref{sec:WMD}, to measure the content shift between the original sentence and the rewritten one. We calculate $\mathrm{WMD}(\bm{x_i}, \bm{\hat y_i})$ and $\mathrm{WMD}(\bm{x_i}, \bm{\mathrm{match}_i})$. If the latter one is smaller, we then replace $\bm{\hat y_i}$ with $\bm{\mathrm{match}_i}$ in $\hat Y^{(t)}$; otherwise, we just keep the original $\bm{\hat y_i}$. In this way, we can obtain an updated version of $\hat Y^{(t)}$ after this matching step.

\paragraph{(Step 2) Translation} In each iteration $t\geq 0$, a Seq2Seq machine translation model with attention $M^{(t)}$ \cite{luong2015effective} is trained from scratch using the pseudo-parallel corpus $\hat X$ and $\hat Y^{(t)}$.

\paragraph{(Step 3) Refinement} In this step, we refine the $\hat Y^{(t)}$ obtained in step 1 with the trained translation model $M^{(t)}$ in step 2. Specifically, we apply the model $M^{(t)}$ to translate each sentence $\bm{x_i}\in \hat X$ to $\bm{\mathrm{trans}_i}^{(t)}$, and form a \textit{temporary} corpus $\mathrm{Trans}^{(t)}$. Again, for any sentence $\bm{x_i}\in \hat X$, we now have two pseudo-parallel sentences $\bm{\hat y_i}\in \hat Y^{(t)}$ and $\bm{\mathrm{trans}_i}\in \mathrm{Trans}^{(t)}$, one of which is from the matching step and the other is from the translation model. We still compare $\mathrm{WMD}(\bm{x_i}, \bm{\hat y_i})$ with $\mathrm{WMD}(\bm{x_i}, \bm{\mathrm{trans}_i})$, and choose the sentence with the smaller value and insert into $\hat Y^{(t+1)}$, which will be fed into step 1 of the next iteration. 

Overall, the aforementioned three steps are repeated for several iterations. This process is formalized below in Algorithm \ref{alg:transfer}.

% \vfill\null

% \begin{figure}
% \Centering
% \begin{multicols}{2}

% \begin{figure}[htb]
%   \centering
%   \begin{minipage}{.7\linewidth}

\begin{algorithm}[ht]
\small
\caption{Iterative Matching and Translation}
\label{alg:transfer}

\begin{algorithmic}
\REQUIRE Two corpora $X, Y$ with different attributes. 

\STATE $\hat X = \{\}$
\FOR{$t=0,\cdots,T$}
    \STATE $\hat Y^{(t)} = \{\}$
    
    \FOR{$\bm{x_i}\in X$}
        \IF{ $t == 0$}
        
        \IF{$\max_{\bm{y} \in Y} \mathrm{Sim}(\bm{x_i}, \bm{y}) > \gamma$ }
        	\STATE $\bm{\hat y_i} = \argmax\limits_{\bm{y} \in Y} \mathrm{Sim}(\bm{x_i}, \bm{y})$
            \STATE $\hat X$.append($\bm{x_i}$)
            \STATE $\hat Y^{(0)}$.append($\bm{\hat y_i}$)
            % \STATE $\hat Y^{(0)}$.append($\bm{\hat y_i}$)
        \ENDIF
        \ELSE
            \STATE $ \bm{\mathrm{match}_i} \leftarrow \argmax\limits_{\bm{y} \in Y}\mathrm{Sim}( \bm{\hat y_i},\bm{y})$
            \STATE $\bm{\hat y_i} \leftarrow \argmin\limits_{\bm{y} \in {\{\bm{\hat y_i} , \bm{\mathrm{match}_i} \}}}\mathrm{WMD}(\bm{x_i}, \bm{y})$
            \STATE $\hat Y^{(t)}$.append($\bm{\hat y_i}$)
        \ENDIF
    \ENDFOR

\STATE
\STATE Train a Seq2Seq model $M^{(t)}:\hat X \rightarrow \hat Y^{(t)}$
\FOR{$\bm{x_i}\in \hat X$}
    
	\STATE $ \bm{\mathrm{trans}_i} \leftarrow M^{(t)}(\bm{x_i})$
	\STATE $ \bm{\hat y_i} \leftarrow \argmin\limits_{\bm{y} \in {\{\bm{\hat y_i}, \bm{\mathrm{trans}_i}\} }}\mathrm{WMD}(\bm{x_i}, \bm{y})$
	\STATE $\bm{\hat y_i}^{(t)} \leftarrow \bm{\hat y_i}$
	
\ENDFOR
\ENDFOR
\ENSURE Attribute transfer model $M^{(T)}: X \rightarrow \hat Y^{(T)}$
\end{algorithmic}

\end{algorithm}
% \end{minipage}
% \end{figure}
% \end{multicols}
% \end{figure}

\subsection{Method Details}

\paragraph{Word Mover Distance}\label{sec:WMD}

In our algorithm, WMD is used to measure the content shift from the source sentence to the rewrite. WMD is a metric of ``travel cost'' from sentence $\bm{s_a}$ to $\bm{s_b}$. The detailed explanations and calculation of the distance is in the paper \cite{kusner2015word}. In brief, each sentence is represented as a weighted point cloud of embedded words. The distance between the sentence $\bm{s_a}$ to $\bm{s_b}$ is the minimum cumulative distance that words from sentence $\bm{s_a}$ need to travel to match exactly the point cloud of sentence $\bm{s_b}$. Denote the vocabulary size as $n$, the travel distance from the word $i$ in sentence $\bm{s_a}$ to the word $j$ in sentence $\bm{s_b}$ as $T(i,j)$, and the corresponding cost of this ``word travel'' as $c(i,j)$. The distance calculation is formulated as 
\begin{align}
    \mathrm{WMD}(\bm{s_a}, \bm{s_b}) = \min_{\mathbf{T} \geq 0}  \sum_{i,j=1}^n \mathbf{T_{ij}} \cdot c(i,j).
\end{align}

Since the initial construction of pseudo-parallel corpus is already guaranteed to have good target attribute and grammaticality, the only remaining criteria to fulfill is the minimal content change from the original sentence to the resulted output. WMD is used as a decision factor whenever an update occurs. Keeping the sentence pair with the smallest cost from each other approximates minimization of the content shift.

Advantages of the WMD over other basic measures of sentence similarity include the fact that it has no hyperparameters to tune, appropriately handles sentences with unequal number of words (via weighted word matchings that sum to 1), accounts for synonymic-similarity at the word-level (through use of pretrained word embedding vectors), and considers the entire contents of each sentence (every word must be somehow matched).  Furthermore, WMD has produced high accuracy in information retrieval \citep{brokos2016using,kim2017bridging}, where measuring content-similarity is critical (as in our attribute transfer task).

\paragraph{Semantic Sentence Representation}

For the matching process in Step 1, the cosine similarity is computed between each pair of sentences. There is no perfect way of semantic representation of a sentence, but a state-of-the-art method is to use the sentence embeddings obtained by averaging the ELMo embeddings of all the words in the sentence \cite{peters2018deep}. ELMo uses a deep, bi-directional LSTM model to create word representations within the context that they are used. \citet{perone2018evaluation} have shown that this approach can efficiently represent semantic and linguistic features of a sentence, outperforming more elaborate approaches such as Skip-Thought~\cite{kiros2015skip}, InferSent~\cite{conneau2017supervised} and Universal Sentence Encoder~\cite{cer2018universal}.

\section{Experiments}

\iffalse 
\subsection{Baselines}
\textit{CrossAlignment} (CA)~\citep{shen2017style} and  \textit{Multi-Decoder} (MD)~\citep{fu2017style} use adversarial methods to encode a sentence into a style-independent latent content vector, which is then fed into a style-dependent decoder for transfer. 
\textit{DeleteAndRetrieve} (DAR)~\citep{li2018delete} first extracts content words by deleting phrases marked as ``sentiment-distinctive'', then retrieves new phrases associated with the target sentiment, and uses a neural model to combine them into a final output.
% This simple method is very effective and outperforms adversarial approaches.
%that struggle to produce high-quality outputs.

\subsection{Experimental Details}
We use 1-layer LSTMs with 1024 hidden dimensions for the encoder and decoder of the sequence-to-sequence translation model with attention. More experimental details are included in Appendix~\ref{sec:appendix_exp_details}.
\fi 

\subsection{Datasets} \label{sec_dataset}
We evaluate the proposed model on two representative tasks: sentiment modification on the \textsc{Yelp} dataset, and text formality conversion on the \textsc{Formality} dataset. A careful human assessment in Section~\ref{dataset_quality} shows that these two datasets are significantly more suitable than the other three popular ``style transfer'' datasets, namely the political slant transfer dataset~\cite{prabhumoye2018style, tian2018structured}, gender transfer dataset~\cite{prabhumoye2018style}, and humorous-to-romantic transfer dataset~\cite{li2018delete}.

\begin{table}[ht]
  \small
  \begin{center}
    \begin{tabular}{c|l|cccc} 
    \toprule
      Dataset & Style & Train & Dev & Test \\ \hline
      \multirow{ 2}{*}{\textsc{Yelp}} &Positive&  270K  & 2K & 0.5K \\
      &Negative & 180K & 2K & 0.5K  \\
      \hline
      \multirow{ 2}{*}{\textsc{Formality}} &Formal& 90K &  4.6K & 2.7K  \\
      &Informal& 96K  &  5.6K & 2.1K \\ 
      \toprule
    \end{tabular}
    \caption{Statistics of \textsc{Yelp} and \textsc{Formality} datasets.}
    \label{tab:dataset_stats}
  \end{center}
\end{table}
\paragraph{\textsc{Yelp}} The commonly used \textsc{Yelp} review dataset for sentiment modification~\citep{shen2017style,li2018delete,prabhumoye2018style} contains a positive corpus of reviews rated above three and a negative corpus of reviews rated below three. This task requires flipping high polarity sentences such as ``The food is good'' and ``The food is bad''.
We use the same train/test split as~\citet{shen2017style, li2018delete} (see Table~\ref{tab:dataset_stats}). 

\paragraph{\textsc{Formality}} The \textsc{Formality} dataset stems from an aligned set of formal and informal sentences \cite{rao2018dear}. It demands changes in subtle linguistic features such as ``want to'' to ``wanna''. We obtained a non-parallel dataset by shuffling the two originally aligned corpora (removing duplicates and sentences that exceed 100 words). Table~\ref{tab:dataset_stats} describes statistics of the resulting two corpora. The development/test sets are  provided with four human-generated attribute transfer rewrites for each sentence (i.e.\ the gold standard).

\subsubsection{Dataset Quality Assessment}
\label{dataset_quality}In order to identify the best datasets for evaluation, we asked human judges to assess five popular text attribute rewriting datasets, namely \textsc{Yelp} sentiment modification dataset, \textsc{Formality} dataset, \textsc{Political} slant transfer dataset~\cite{prabhumoye2018style, tian2018structured}, \textsc{Gender} transfer dataset~\cite{prabhumoye2018style}, and \textsc{Humorous}-to-\textsc{Romantic} transfer dataset~\cite{li2018delete}. For each dataset, we randomly extracted 100 sentences (i.e. 50 per style). For each sentence, we asked two native English speakers to annotate them with one of three options, i.e. either of the two attributes or ``Cannot Decide''. Based on the collected annotations, we calculate three metrics: (1) \textit{undecidable rate}, which is the percentage of ``Cannot Decide'' answers (we report the average percentage between the two annotators), (2) \textit{disagreement rate}, which is the percentage of different opinions between the two annotators, and (3) \textit{F1 score} between the human annotations and the gold labels in the original dataset (``Cannot Decide'' answers were not considered).

\begin{table}[ht]
\centering
\small
\begin{tabular}{l|ccc}
\toprule
Dataset   & Unde. Rate & Dis. Rate & F1 \\ \hline
\textsc{Political} & 49.0                  & 33.0                  & 67.2    \\ 
\textsc{Gender}    & 88.0                  & 14.0                  & 53.6    \\
\textsc{Humorous}  & 66.0                  & 31.0                  & 79.3    \\\hline
\textsc{Yelp}      & \textbf{11.0}                  & \underline{16.0}                  & \textbf{93.2}    \\
\textsc{Formality} & \textbf{0.5}                   & \underline{17.0}                  & \textbf{87.6}   \\ 
\toprule
\end{tabular}
\caption{Comparison of dataset quality. ``Unde.\ Rate'' indicates the undecidable rate, ``Dis.\ Rate'' the disagreement rate between annotators.}
\label{table:dataset-comparison}
\end{table}

\begin{table*}[!ht]
  \small
  \begin{center}
    \begin{tabular}{l|cccc|cccc} 
    \toprule

	   & \multicolumn{4}{c|}{\textsc{Yelp}}& \multicolumn{4}{c}{\textsc{Formality}} \\ 
       & Content & Grammar & Attribute & \textit{Success (\%)}  & Content & Grammar & Attribute & \textit{Success (\%)} \\ \hline
      {CA}  &  2.30 \tiny $\pm 1.45$  & 2.65 \tiny $\pm 1.32$ & 2.77 \tiny $\pm 1.69$ & 5.5 & 2.45 \tiny $\pm 1.31$ & 2.67 \tiny $\pm 1.57$  & 1.97 \tiny $\pm 1.26$ & 0.5\\
      {MD}  & 2.12 \tiny $\pm 1.52$ & 2.38 \tiny $\pm 1.49$ & 2.25 \tiny $\pm 1.54$ & 2.0 & 1.79 \tiny $\pm 1.17$ & 2.37 \tiny $\pm 1.50$ & 2.42 \tiny $\pm 1.56$ & 0.0 \\
      {DAR}  & 2.93 \tiny $\pm 1.45$  & 2.95 \tiny $\pm 1.51$ & 3.09 \tiny $\pm 1.65$ & 14.5 & 2.89 \tiny $\pm 1.39$ & 3.02 \tiny $\pm 1.59$  & 2.95 \tiny $\pm 1.56$ & 7.0\\
      \hline
      Ours  & \textbf{3.07} \tiny $\pm 1.49$ &  \textbf{4.32} \tiny $\pm 1.07$ & \textbf{3.43} \tiny $\pm 1.65$ & \textbf{21.0} & \textbf{3.60} \tiny $\pm 1.55$ & \textbf{4.42} \tiny $\pm 1.08$ & \textbf{3.11} \tiny $\pm 1.58$ & \textbf{29.0} \\ 
      \hline
      Human  & 4.66 \tiny $\pm 0.77$  & 4.33 \tiny $\pm 1.07$ & 4.22 \tiny $\pm 1.29$ & 63.5 & 4.62 \tiny $\pm 0.82$ & 4.62 \tiny $\pm 0.94$ & 3.91 \tiny $\pm 1.57$ & 59.5\\ 
      \toprule
    \end{tabular}
    \caption{The mean and standard deviation of human ratings on Content preservation, Grammaticality, and attribute correctness on both datasets for different systems: \textsc{\textbf{C}ross\textbf{A}lignment} (CA), \textsc{\textbf{M}ulti\textbf{D}ecoder} (MD), \textsc{\textbf{D}elete\textbf{A}nd\textbf{R}etrieve} (DAR), \textbf{Ours}, and \textbf{Human} reference. Ratings are on a 1 to 5 scale. }
    \label{tab:human-eval}
  \end{center}
\end{table*}
The scores for each dataset are summarized in Table \ref{table:dataset-comparison}. A quick comparison shows that \textsc{Yelp} and \textsc{Formality} obtain significantly lower \textit{undecidable} and inter-annotator \textit{disagreement rates}, indicating that the style of the sentences in these two datasets are less ambiguous to humans. In addition, \textsc{Yelp} and \textsc{Formality} have much higher F1 score than all the other datasets, which confirms the correctness of the source-target attribute split. % \myworries{Is there any interannotator agreement? you may calculate cohen or fleiss eventually...} \note{the disagreement rate is inter-annotator. Does it look clear now?}

This comparison points out that the three datasets, including political slant, gender and romantic-to-humorous caption datasets, are ambiguous and noisy, therefore not only adding complexity to the task and its evaluation (because even human annotators struggle to identify the correct style), but also leading to models that are not useful in real practice. %.\newadd{Furthermore, a system trained on such noisy data cannot be applicable and useful.}%\myworries{ok, but more importantly, a system that is trained on such a noisy data is useless. make this paragraph sharper}

\subsection{Evaluation Strategies} \label{sec:eval_hum_auto}
\paragraph{Human Evaluation} Following \citet{li2018delete, lample2018multipleattribute}, we asked human judges to evaluate outputs of different models in terms of content preservation, grammaticality and attribute change correctness on a Likert scale from 1 to 5. We randomly selected 100 sentences from each corpus (100 positive and 100 negative sentences from \textsc{Yelp}; 100 formal and 100 informal sentences from \textsc{Formality}). % \myworries{It is a bit unclear whether these selected sentences are outputs or not, as you speak about corpus... Can you clarify} 
Each of the twelve human judges passed a test batch prequalification before they could start evaluating, and we verified they spent a reasonable amount of time in the task.

\paragraph{Automatic Evaluation}
In addition to the human evaluation, we also programatically gauge rewriting quality via automated methods, following the practice in \cite{li2018delete, lample2018multipleattribute}. Here, content preservation is measured by the BLEU score between model outputs and multiple human references for each sentence.\footnote{We used the script to calculate BLEU after detokenization at \url{https://github.com/moses-smt/mosesdecoder/blob/master/scripts/generic/multi-bleu-detok.perl.}} Since the previous work only introduces single human reference for \textsc{Yelp}, we enriched the test set with four extra human references for each test sentence, in total 800 rewrites. The diverse human rewrites ensures a more variety-tolerant measurement of model outputs. Attribute accuracy is assessed via the classification results of a CNN-based text classifier~\cite{kim2014convolutional}. To automatically gauge fluency, we use pretrained language models (LM) to get the perplexity of output sentences.
The details of the classifier, LM and the collection of human references are elaborated in Appendix~\ref{sec:appendix_auto_setup}.

% We measure the performance of our system by two automatic metrics: classifier-based sentiment correctness and BLEU score~\citep{papineni2002bleu} with respect to human references. The classifier is trained on the original \textsc{Yelp} dataset to predict the rating for reviews within 5 sentences. The percentage of model outputs labeled with the expected target sentiment by the classifier is used to measure the style correctness.

% To analyze the content preservation and fluency, we compute the BLEU score between model outputs and five human transfers. The five human references transfer the test sentences in different ways so a comprehensive BLEU score among all five gold outputs represents the quality of the outputs.

% Shen BLEU: 3.1
% Li BLEU: 8.47
% Our BLEU: 8.22 (this is the bleu of 1 reference, I need to update this score on 5 references --Zhijing)

% Shen sentiment accuracy: 73.7%
% Li sentiment accuracy: 88.22%，
% Our sentiment accuracy: 99%

% We tested the proposed style transfer approach on two subtasks: flipping the sentiment of \textsc{Yelp} review dataset~\cite{shen2017style, li2018delete, prabhumoye2018style, fu2017style} and changing the text formality~\cite{rao2018dear}. Performance was evaluated both with human judgments and automated metrics. 

\subsection{Baselines}
To compare against multiple baselines, we re-implemented three recently-proposed methods (described in Sections \ref{sec:intro} and \ref{sec:relatedwork}): \textsc{CrossAlignment} (CA) from \citet{shen2017style}, \textsc{MultiDecoder} (MD) from \citet{fu2017style}, and \textsc{DeleteAndRetrieve} (DAR) from \citet{li2018delete}.

\subsection{Experimental settings}
\label{sec:experiments-setting}
For the translation process, we used an off-the-shelf Seq2Seq model, a 2-layer LSTM encoder-decoder with 1024 hidden dimensions and attention. We focused on the novelty of the proposed iterative framework, so a standard and classic Seq2Seq model is used.
More experimental details are described in Appendix~\ref{sec:appendix_exp_details}.

\begin{table*}[!ht]
\small
\centering
\begin{tabular}{>{\raggedright}p{0.0005\textwidth}>{\raggedright}p{0.1\textwidth}>{\raggedright}p{0.5\textwidth}}
\toprule 
 \multicolumn{3}{c}{ \textbf{From {\color{blue}Positive} to {\color{red}Negative} (\textbf{\textsc{Yelp}})} }  \tabularnewline \midrule 
& \textbf{Input} & Thank you ladies for being {\color{blue}\textbf{awesome}}!  \tabularnewline \midrule
& CA & Thank you charge up for this food business.
\tabularnewline
& MD & Thank you for it \$ {\color{red}\textbf{poor}} company.
\tabularnewline
& DAR & Thank you ladies for being {\color{red}\textbf{didn't}}.
\tabularnewline
& \textbf{Ours} & \textbf{Thank you for {\color{red}wasting} my time you {\color{red}idiots}.}
\tabularnewline
% \hline
& Human & Thank you ladies for being the {\color{red}\textbf{worst}}!
\tabularnewline \midrule \midrule 
\multicolumn{3}{c}{ \textbf{From \textcolor{myorange}{Informal} to \textcolor{mygreen}{Formal} (\textbf{\textsc{Formality}})} } \tabularnewline
\midrule
& \textbf{Input} & \textcolor{myorange}{\textbf{i}} tried to like him, bu \textcolor{myorange}{\textbf{i just can't}}.  \tabularnewline
\midrule
& CA & I was to him, but I \textcolor{myorange}{\textbf{just like I can't}}.	\tabularnewline 
& MD & \textcolor{myorange}{\textbf{i}} tried to him like, but I \textcolor{myorange}{\textbf{just should too}}. \tabularnewline 
& DAR & I do not know if you like him but I do not know he \textcolor{myorange}{\textbf{can't}}.	\tabularnewline
& \textbf{Ours} & \textbf{I tried to like him, but I \textcolor{mygreen}{cannot}.} \tabularnewline
% \hline
& Human & I \textcolor{mygreen}{\textbf{attempted}} to like him, \textcolor{mygreen}{\textbf{however}}, I am \textcolor{mygreen}{\textbf{unable}}.
\tabularnewline
\bottomrule
\end{tabular}
\caption{Example outputs of different systems.
}
\label{tab:modeloutputs}
\end{table*}
\section{Results}
\label{sec:results}

\iffalse
We evaluate our methods on the benchmark dataset of \textsc{Yelp} reviews and \textsc{Formality} by both automatic and human evaluation. Our method shows significant improvement over the previous models on both metrics. 
\fi

\subsection{Human Evaluation}
\label{sec_human_results}

In terms of human evaluation, the proposed approach shows significant gains over all baselines in terms of attribute correctness, content preservation, grammaticality, and success rate as shown in Table~\ref{tab:human-eval} ($p < 0.05$ using bootstrap resampling \cite{Koehn2004StatisticalST}). 

The largest improvement is in grammaticality, where we achieve an average of $4.32$ on \textsc{Yelp} and $4.42$ out of $5.0$ on the \textsc{Formality} dataset. These scores are close to those of human references, prevalently outperforming the baselines. On attribute correctness, the model scores $3.43$ on \textsc{Yelp} and $3.11$ on \textsc{Formality} dataset, exceeding the previous best methods by $0.33$ and $0.16$ respectively. Moreover, in content preservation, we show an improvement of $0.14$ and $0.72$ over the previous state-of-the-art \textit{DeleteAndRetrieve} model on the two datasets. Finally, we follow the practice of \citet{li2018delete} to evaluate the \textit{Success Rate} --- an aggregate of the three previous metrics, in which a sample is successful only if it is rated $4$ or $5$ on all the three metrics. Our model demonstrates an improvement of $6.5\%$ in success rate over the previous best method. %, where our model outscores the baselines by about one half on \textsc{Yelp} dataset and even more than three times on \textsc{Formality} dataset.

% Overall, the proposed iterative matching and translation algorithm starts from a attribute-perfect iteration zero, as all the target sentences are directly from the desired attribute $a_2$, and then increases the content preservation through the translation and iterative process. 

To further inspect performance of different methods, we show some typical outputs in Table~\ref{tab:modeloutputs}. First, the output of our model is nearly as grammatical as the human-written sentence, compared with the loss of sentence structure in \textsc{CrossAlignment} (e.g.\ ``I tried to him like'') and  \textsc{DeleteAndRetrieve} (e.g.\ ``for being didn't''). Second, in terms of attribute correctness, \textsc{DeleteAndRetrieve} suffers from failure to convert the attribute-bearing words (as it wrongly converts the word ``awesome'' to ``didn't''), and \textsc{CrossAlignment} is prone to missing attribute words. Third, although our method enforces content alignment, the preservation of content is an inevitable challenge for the previous approaches such as \textsc{MultiDecoder} and \textsc{DeleteAndRetrieve}.

\subsection{Automatic Evaluation} \label{sec_auto_results}

\begin{table}[t]
\small
  \begin{center}
      \begin{tabular}
      {p{.083\columnwidth}|p{.09\columnwidth}p{.09\columnwidth}p{.09\columnwidth}|p{.09\columnwidth}p{.09\columnwidth}p{.09\columnwidth}} 
        \toprule
    
    	   & \multicolumn{3}{c|}{\textsc{Yelp}}& \multicolumn{3}{c}{\textsc{Formality}} \\ 
           & Acc. & BLEU & PPL & Acc. & BLEU & PPL \\ \hline
          CA &73.40 & 17.54  & 46.92  & 46.41& 12.65 & 53.89 \\
        %   Fader  &  \textbf{9.56}& \textbf{46.59} &   &  & \\
          MD  &66.80 & 17.34 & 69.51  & 66.46& 7.46 & 109.41 \\
          DAR & 88.00 &\textbf{31.07} & 79.70 & \textbf{78.87}&  21.29  & 63.61 \\ \hline
          Ours & \textbf{95.90} & 22.46 & \textbf{14.89} & 72.07& \textbf{38.16}  & \textbf{32.63}\\ 
        %   \hline
        %   Human  &  &  &  &  & \\ 
          \toprule
        \end{tabular}
        \caption{Automatic evaluation results on both datasets. Acc. is the style/attribute classification accuracy (\%).}
        \label{tab:auto-eval}
    
  \end{center}
\end{table}

% Following previous work~\cite{shen2017style, fu2017style, li2018delete, prabhumoye2018style}, 

We also assess each method's performance via the automatic measures of attribute-accuracy, BLEU, and perplexity (Table~\ref{tab:auto-eval}).
% The automatic evaluation shows a rough comparison of all model outputs.
% Although the results are not perfectly comparable to the more reliable ground truth rating provided by human judges (Table~\ref{tab:human-eval} and Table~\ref{tab:auto-eval}), they serve as a convenient indicator when tuning the models or evaluating some intermediate outputs. 
A highlight is its perplexity score, outperforming the previous methods by a large margin. This advantage owes to the fact that the Seq2Seq model is trained on pseudo-pairs similar to real samples, which can guarantee the translation quality. However, a tradeoff between the other two aspects, attribute accuracy and BLEU, can be clearly seen on both \textsc{Yelp} and \textsc{Formality} datasets. This is common when tuning all models, as targeting at a higher BLEU score will result in a lower attribute correctness score~\cite{shen2017style, fu2017style, li2018delete}. Similarly, in iterations of our model, the BLEU score gradually increases while the accuracy decreases. Therefore, the reported outputs are balanced based on all three aspects. An analysis on the limitations of automatic evaluation is in Appendix~\ref{sec:appendix_auto_limit}.

% \vspace{0.5mm}
\section{Performance Analysis}

Here, we analyze the performance of our model with regard to various aspects in order to understand what factors underlie its success.
% with respect to the effective initialization of the pseudo-parallel corpus, the denoising translation process, and the iterative refinement. % \myworries{Check a bit the paper and avoid repeating too many times stuff like 'further investigate the strengh of our approach...'. this kind of wording is ok only in the abstract, introduction and conclusions. in the rest of the paper, you should speak about 'output of the model', keeping a more objective wording; otherwise you will sound bias and redundant}

\subsection{Initialization of Pseudo-Parallel Corpus}
\label{sec:analysis-matching}
The proposed model starts from the construction of an initial pseudo-parallel corpus by our matching step. Note that this initial pairing is practicable in most domains. For example, Yelp reviews naturally have positive and negative comments on the same food; different translations of the same book also shares the same content; different news agencies report with different tones on the same events. After construction, there are three properties of this pseudo-parallel corpus: First, it is a subset of the original corpus.
% (100\% and 85\% percent of sentences are matched for the \textsc{Formality} and \textsc{Yelp} datasets, respectively)
Second, all retrieved target sentences contain the desired attribute information and are of perfect grammaticality. This property is retained throughout our iterations and is key to the high attribute transfer accuracy and fluent language generation capabilities of our model.
% as we observed the attribute accuracy and perplexity do not change much over the iterations.
Third, the sentence pairs are 
% of reasonable quality but 
still imperfect in terms of content preservation, often similar in meaning but with certain content words altered. This is remedied by subsequent refinement steps.
% However, the initial semantic sentence matching of ELMo embeddings still provides reasonable quality to as the later iterative process can achieve good results. 
% It is difficult to get an exact quality threshold, so we try to inspect the lower bound of the corpus quality that enables the working of the model, providing convenience for future investigation. 

Although our model needs the matched pseudo-parallel corpus as a starting point, it has high tolerance to recover from occasional low-quality matches. 
% In order to inspect the quality of our initial pseudo-parallel corpus,
To demonstrate this, we randomly picked 100 sentences and their initial pseudo-pairs from both source and target corpus, and asked human judges to rate them. For each sentence pair, three judges decided whether the sentence pair forms a good rewrite, a bad rewrite, or an ambiguous one. We mark the pseudo-pair as either good or bad if at least two annotators agree on such a judgment. The percentage of bad rewrites is $38.2\%$ and $48.2\%$ on \textsc{Yelp} and \textsc{Formality}, respectively. This indicates that subsequent iterative refinement indeed allows for $30-50\%$ low-quality pairs in the initial pseudo-parallel corpus.

\subsection{Effective Denoising Translation}
% Based on the high style accuracy and grammaticality given by the initial pseudo-parallel corpus, 
One of the most important gains from the iterative translation is to encourage more content preservation. We illustrate the effectiveness of translation by using automatic evaluation to gauge how the bad matches in the initial pseudo-parallel corpus change \textit{immediately after} the first translation step. We find that after the first translation, the BLEU score of these bad matches shows a clear improvement, increasing from 9.40 to 13.18 on \textsc{Yelp}, and 5.44 to 28.25 on \textsc{Formality}. This shows that the translation model recognizes the noise in the first matching process and generates more proper transfer candidates, providing a good foundation for subsequent iterative refinement. An illustrative example of this refinement can be seen in the sentence (A) in Figure~\ref{fig:method}, where a bad match of ``Worst burrito ever ever'' was denoised and replaced with ``Worst pizza ever ever'' after the first translation.

\subsection{Iterative Refinement}
\begin{figure}[ht]
\small
\begin{center}
\begin{tikzpicture}
\pgfplotsset{
    scale only axis,
    scaled x ticks=base 10:0,
    xmin=0, xmax=5,
    xtick={0,...,5},
    legend style={at={(0.71,0.3)},anchor=north},
}
\begin{axis}[
  legend style={cells={align=left}},
  axis y line*=left,
  ymin=0, ymax=42,
  xlabel=Iteration,
  ylabel=BLEU,
]
\addplot[smooth,mark=x,blue]
  coordinates{
    (0,14.98)
    (1,17.86)
    (2,19.13)
    (3,19.21)
    (4,19.42)
    (5,22.30)
}; \label{plot_one}
\addlegendentry{plot 1}
\end{axis}
\begin{axis}[  
  grid=both,
  legend style={cells={align=left}},
  axis y line*=right,
 axis x line*=right,
  ymin=0, ymax=42,
  minor tick num=2,
  ylabel={}
]
\addlegendimage{/pgfplots/refstyle=plot_one}\addlegendentry{\textsc{Yelp}}
\addplot[smooth,mark=*,orange]
  coordinates{
    (0,8.13)
    (1,28.11)
    (2,33.59)
    (3,37.16)
    (4,39.08)
}; \addlegendentry{\textsc{Form.}}
\end{axis}
\end{tikzpicture}
\caption{BLEU score evolution across training iterations on \textsc{Yelp} and \textsc{Formality} datasets.}
\label{fig:iterchart}
\end{center}
\small
\end{figure}
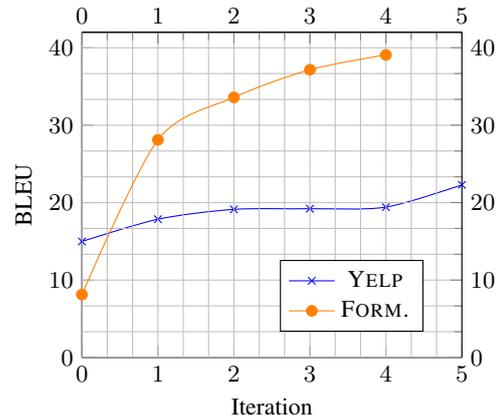

The essential increase of content preservation owes to the iterative refinement step. This process reduces erroneous alignments in the pseudo-parallel corpus by updating each existing pseudo-pair with newly generated translation-pair. Thanks to the denoising effect of the translation model, this updating could improve the quality of pseudo-pairs. Furthermore, we use WMD to measure the content shift between sentences in the pseudo-pair and the translation-pair, and accept the update only when the translation-pair possesses smaller content shift so as to avoid worsening updates. 
% After each iteration, a refined corpus is constructed, facilitating better training for the next round. By iterative alignment and translation, the model gradually learns to identify the content and ignore unmatched parts of the pseudo pairs, obtaining better performance in each round. 
As the iteration goes on, the accuracy and perplexity stay high from the beginning, but as in Figure~\ref{fig:iterchart}, the BLEU score keeps increasing. This shows that the new iteration outputs retains more content from the original sentences. More importantly, this refinement can prevent the model from totally relying on the fine-quality matching pairs, which contributes to the high tolerance capability on the matching quality as discussed in Section \ref{sec:analysis-matching}.
% because the matching and translation ensures all sentences are similar to the target corpus in their attribute and grammaticality. Hence, we compared the only dramatically changing aspect, the BLEU score
% between the source corpus $X$ and the pseudo-parallel corpus $\hat Y^{(t)}$ 
% at the end of each iteration $t$.
% Figure~\ref{fig:iterchart} shows that for both datasets, the BLEU score keeps increasing as the iterative training proceeds, indicating that the outputs are refined after each iteration. 

% Graph for iteration line chart

% \subsection{Composition of Pseudo-Parallel Corpus}

The pseudo-parallel corpus in Iteration ${t\geq0}$ is composed of matched pairs from the original corpora and translated pairs from the trained translation model. Both of them contribute to the good performance of our model. For further investigation, we look into the pseudo-parallel corpus $\hat X$ and $\hat Y^{(4)}$ in the final training iteration for the \textsc{Formality} dataset. Since this dataset has ground-truth parallel pairs, we are able to calculate how many pseudo-parallel pairs are the same as the ground-truth parallel pairs, and the percentage turns out to be around 52\%, which is actually not high. This reveals that the refinement step can provide the model with better pseudo-pairs even than original gold pairs and our model can still 
stand out without fine-quality matches.

\section{Conclusion}

In this work, we proposed a Seq2Seq paradigm for text attribute transfer applications, suggesting a simple but strong method for overcoming lack of parallel data. We construct a pseudo-parallel corpus by iteratively matching and updating in a way that increasingly refines the final transfer function. Our framework can employ any Seq2Seq model and 
% for training a standard sequence-to-sequence translation model, and this process is then iteratively refined.  
%We found that the quality of results increases in each iteration.  Our model substantially 
outperforms previous methods under all measured criteria (content preservation, fluency, and attribute correctness) in both human and automatic evaluation. The simplicity and flexibility of our approach can be useful in applications that require intricate edits or complete sentence rewrites.
% , such as informal to formal style conversion \cite{rao2018dear} or romantic to humorous  \cite{li2018delete}.

\section*{Acknowledgments}

We thank Professor Peter Szolovits for supporting and funding the work, and Prof Benjamin Kao for advice on evaluation metrics. We also appreciate Tianxiao Shen, Jiang Guo, Professor Regina Barzilay, and MIT Natural Language Processing Group for insightful discussions and support. We thank Adam Fisch, Benson Chen, Jingran Zhou, Wui Wang Lui, Shuailin Yuan, Zhutian Yang, Yilun Du and many other helpers for human evaluation.

\bibliography{emnlp-ijcnlp-2019}
\bibliographystyle{acl_natbib}

\clearpage
\appendix

\section{Supplemental Material}
\label{sec:appendix}
In this Appendix, we explain the setup of automatic evaluation, experimental details, and a discussion on limitations of automatic evaluation.

\subsection{Automatic Evaluation Setup}\label{sec:appendix_auto_setup}

The details of the three aspects of automatic evaluation are elaborated as follows.

\begin{itemize}
    \item \textbf{Style correctness:}  Following \citet{shen2017style}, we trained a CNN-based text classifier \cite{kim2014convolutional} on our original datasets, using its accuracy over the system outputs to measure their style correctness. The accuracy of this classifier is respectively 97\% and 93\% on the \textsc{Yelp} and the \textsc{Formality} datasets.
    
    \item \textbf{Content preservation:} To evaluate the content preservation, we compute the BLEU score between model outputs and multiple human references. \textsc{Formality} dataset comes with four human references for around 2,000 formal and 2,000 informal sentences. 
    
    For the \textsc{Yelp} dataset, we used the same test set of 500 positive and 500 negative sentences as \citep{li2018delete}, and collected four references for each sentence in the test set, in addition to the single human reference released by \citet{li2018delete}. We hired crowdworkers on Amazon Mechanical Turk to rewrite the source sentence with the same content but an opposite sentiment. 
    
    These in total five human rewrites of each test sentence ensures a more tolerant measurement of model outputs. We can see a relatively large diversity in human transfers, which is measured by the average BLEU score of one randomly chosen human reference among the other four. The average difference gap between the calculated score, $52.63$, and a perfect BLEU score, $100$, shows that the five human rewritten sentences contain significant lexical differences.  Evaluation based on all five human rewrites thus offers a better measure of transfer quality.
    % \ref{appendix:human-references}
    For example, the two equally acceptable rewrites ``i will definitely not return often!'' and ``I won't be returning any time soon.'' does not have a single word overlap and have a BLEU score $0$ with each other.
     Therefore, five human references enable a more comprehensive evaluation, allowing multiple ways to transfer a sentence.
    % We noticed, in fact, that humans tend to rewrite the test sentences in a different way (e.g. the BLEU score of a random human reference against the others is only 64). Details of human references collection are reported in the Appendix \ref{appendix:human-references}.
    \item \textbf{Fluency:} Fluency is measured by the perplexity (PPL) of the generated outputs by a pre-trained language model (LM) using GluonNLP toolkit\footnote{https://gluon-nlp.mxnet.io/index.html}. The encoder of this LM is comprised of two long-short term memory (LSTM) layers, each of which has 200 hidden units. The embedding and output weights are tied. Dropout of 0.2 was applied to both embedding and LSTM layers. LM was optimized via stochastic gradient descent (SGD) optimizer with learning rate of 20 for 15 epochs. 
    % \ref{appendix:ppl}
    
\end{itemize}

\subsection{Experimental Details}\label{sec:appendix_exp_details}

We adopt the 100-dimensional pretrained GloVe word embeddings~\citep{pennington2014glove} as inputs of a standard machine translation sequence-to-sequence model with attention. The NLTK software package is used to generate Part-of-Speech tags and we feed them as additional input to the encoder. To ensure both a relatively high quality pseudo-parallel corpus and no significant drop in size on the two corpora, we only match sentences with vector cosine higher than an empirical similarity threshold of $0.7$. We train the model until the update rate of candidate transfer is lower than $0.5\%$. This convergence is at iteration $T = 5$ for \textsc{Yelp} and $T = 4$ for \textsc{Formality}. 

\subsection{Limitations of Automatic Evaluation}\label{sec:appendix_auto_limit}

Evaluating the quality of a transferred sentence by a pretrained classifier (style accuracy), lexicon overlap with references (BLEU), and a pretrained Language Model (perplexity) can have many limitations. First, in terms of style accuracy and perplexity, the pretrained models can be unreliable when evaluating a sentence different from the training corpus. For example, comparing human-rated and machine-evaluated score of \textit{style correctness} (Table~\ref{tab:human-eval} and Table~\ref{tab:auto-eval}), we find that although the automatic score can serve as a rough assessment of the models, it does not aligns perfectly with the human ratings in Table~\ref{tab:human-eval}. As is explained in \citep{li2018delete}'s work, the uneven distribution of some content words in two style corpora may confuse the classifier and make it overfitted on the training data. 
% Similarly, the pretrained language models may have the same limitations when evaluating the fluency of outputs.

Second, the BLEU score, which mainly relies on the lexicon overlap between evaluated sentence and references, can mistakenly favor sentences with a higher similarity to the source sentence. An illustration is that by simply copying all source sentences in the test set we can get a BLEU score of 62, despite an accuracy score close to zero. The state-of-the-art model, \textit{DeleteAndRetrieve}, only modifies a few attribute-carrying words and copies the rest of the sentence. For example, it transfers the source sentence ``My `hot' sub \textit{was cold} and the meat was watery.'' into an ungrammatical one ``My `hot' \textit{is a great place} to the meat.'' but keeps the words ``hot'' and ``meat''. Consequently, it results in a high BLEU score and a poor perplexity score.

\end{document}